\pdfoutput=1
\documentclass[nonblindrev]{informs3_arXiv}

\OneAndAHalfSpacedXI 

\usepackage[draft]{hyperref}
\RequirePackage{fix-cm} 
\usepackage{xspace}
	\newcommand\ie{i.\,e.\xspace}
	\newcommand\eg{e.\,g.\xspace}

	\usepackage{siunitx}
    \DeclareSIUnit\eur{\officialeuro}
    \DeclareSIUnit\M{M}
    \DeclareSIUnit\k{k}
	
	\usepackage[%
  sort&compress
]{cleveref}
  \crefname{chapter}{section}{sections}
	\Crefname{chapter}{Section}{Sections}

\usepackage{etoolbox}
\robustify\bfseries

\usepackage{url}

\usepackage{isomath}
\usepackage{bm}

\usepackage{colortbl}
\usepackage{tabularx}
\usepackage{booktabs}
\usepackage{collcell}
\usepackage{subfig}

\usepackage{float}
\usepackage{rotating}

\usepackage{array}
\newcolumntype{L}[1]{>{\raggedright\let\newline\\\arraybackslash\hspace{0pt}}p{#1}}
\newcolumntype{C}[1]{>{\centering\let\newline\\\arraybackslash\hspace{0pt}}p{#1}}
\newcolumntype{R}[1]{>{\raggedleft\let\newline\\\arraybackslash\hspace{0pt}}p{#1}}

\usepackage{csquotes}
\usepackage{amsmath}
\usepackage{amssymb}
\usepackage{pifont}

\usepackage[
    colorinlistoftodos,
    textsize=footnotesize,
        ]{todonotes}

\makeatletter
    \renewcommand{\fps@figure}{H}         
    \renewcommand{\fps@table}{H}         
\makeatother 

\usepackage{float}
\usepackage{rotating}

\makeatletter
\newcolumntype{B}[3]{>{\boldmath\DC@{#1}{#2}{#3}}c<{\DC@end}}
\makeatother

\usepackage{natbib}
 \bibpunct[, ]{(}{)}{,}{a}{}{,}%
 \def\bibfont{\small}%
\TheoremsNumberedThrough     

\EquationsNumberedThrough    

\begin{document}


 \RUNAUTHOR{Lutz, Pr\"ollochs, and Neumann}

\RUNTITLE{Two-Sided Argumentation in Online Consumer Reviews}

\TITLE{Understanding the Role of Two-Sided Argumentation in Online Consumer Reviews: A Language-Based Perspective}

\ARTICLEAUTHORS{%
\AUTHOR{Bernhard Lutz}
\AFF{University of Freiburg, \EMAIL{bernhard.lutz@is.uni-freiburg.de}, \URL{}}
\AUTHOR{Nicolas Pr\"ollochs}
\AFF{University of Oxford, \EMAIL{nicolas.prollochs@eng.ox.ac.uk} \URL{}}
\AUTHOR{Dirk Neumann}
\AFF{University of Freiburg, \EMAIL{dirk.neumann@is.uni-freiburg.de} \URL{}}
} 

\ABSTRACT{%
This paper examines the effect of two-sided argumentation on the perceived helpfulness of online consumer reviews. In contrast to previous works, our analysis thereby sheds light on the reception of reviews from a language-based perspective. For this purpose, we propose an intriguing text analysis approach based on distributed text representations and multi-instance learning to operationalize the two-sidedness of argumentation in review texts. A subsequent empirical analysis using a large corpus of Amazon reviews suggests that two-sided argumentation in reviews significantly increases their helpfulness. We find this effect to be stronger for positive reviews than for negative reviews, whereas a higher degree of emotional language weakens the effect. Our findings have immediate implications for retailer platforms, which can utilize our results to optimize their customer feedback system and to present more useful product reviews. 
}


\HISTORY{}

\maketitle

\section{Introduction}
\label{sec:introduction}


Online consumer reviews represent a key source of information for customers considering purchasing a product \cite[\eg][]{Dellarocas.2003}. 
On modern online retailer platforms, users are typically provided with the opportunity to assign a product to a star rating ranging from one star (very negative) to five stars (very positive). These product valuations not only inform other customers about the quality of a product, but also have a significant and positive impact on retail sales \citep{Chevalier.2006}. Another relevant feature of modern retailer platforms is that customers are typically provided with the opportunity to rate the perceived helpfulness of a review, \ie the extent to which it facilitates their decision-making \citep{Mudambi.2010}. 
Existing research has demonstrated that reviews that are perceived as more helpful also have a greater influence on retailer sales \citep{Dhanasobhon.2007}. Since review helpfulness serves as focal point to study human decision-making, several studies have focused on the question of what makes reviews helpful or unhelpful. For instance, longer and more detailed reviews are perceived as more helpful \citep{Yin.2016}. Previous literature has, however, produced mixed results regarding the effect of review ratings on helpfulness. 
While, for example, \citet{Sen.2007} associate a greater helpfulness of negative reviews, the results from \citet{Mudambi.2010} point in the opposite direction.


Apart from the numeric star ratings, customer reviews typically also contain a substantial amount of unstructured textual data, \ie the review texts. This written content encompasses highly customer-relevant information, such as user experiences or customer opinions \citep{Cao.2011}. Nonetheless, previous works have primarily studied review helpfulness on the basis of structured information (such as star ratings or review length), whereas the textual component has been largely ignored. This stems from the fact that language offers a rich source of information, with direct effects on human decision-making \citep{Pennebaker.2003}. In this context, a particularly decisive aspect is the two-sidedness of argumentation, \ie that the reviewer illustrates both the positive and negative aspects of a particular product. For example, marketing research suggests that \emph{\enquote{two-sided messages generate relatively high levels of attention and motivation to process because they are novel, interesting, and credible}} \citep{Crowley.1994}. 
Since online product reviews, similar to marketing campaigns, advocate a certain opinion about a product, one could expect two-sided argumentation in customer reviews to play an important role regarding their helpfulness.


In order to address this important research question, this paper examines the effect of two-sided argumentation on the perceived helpfulness of customer reviews. For this purpose, we use a dataset of \emph{192,189} Amazon customer reviews in combination with a novel text analysis method that allows us to study the line of argumentation on the basis of individual sentences. As detailed later in this paper, the method employs \emph{distributed text representations} and \emph{multi-instance learning} to transfer information from the document level to the sentence level. By assigning similar sentences to the same polarity label and differing sentences to an opposite polarity label, we are able to operationalize the two-sidedness of argumentation from a language-based perspective. 
A subsequent empirical analysis suggests that two-sided argumentation in reviews significantly increases the helpfulness of the reviews. Moreover, we find this effect to be stronger for positive reviews than for negative reviews, whereas a higher degree of emotional language weakens the effect.


This work immediately suggests manifold implications for practitioners and Information Systems research: we present a language-based approach to better understanding the role of two-sided argumentation in the assessment of customer reviews. In a next step, this allows practitioners to enhance their communication strategies with respect to product descriptions, social media content, and advertising. Moreover, our findings have immediate implications for retailer platforms, which can utilize our results to optimize their customer feedback system and present more useful product reviews. Ultimately, this study contributes to IS research by addressing the paramount question of how humans react to information in the form of written text. 


The remainder of this work is structured as follows. \Cref{sec:background} establishes the background of our study and derives our research hypotheses. In \Cref{sec:methodology}, we introduce our research methodology. 
Subsequently, \Cref{sec:results} presents our empirical setup and tests our hypotheses. In \Cref{sec:discussion} we discuss implications of our findings. Finally, \Cref{sec:conclusion} concludes and outlines our further research agenda.

\section{Research Hypotheses}
\label{sec:background}


In this study, we aim to understand the role of two-sidedness in the helpfulness of customer reviews. Existing research in this direction has focused on the role of review extremity, \ie whether the review rating is positive, negative, or neutral. 
For example, \citet{Pavlou.2006} found that the extreme ratings of sellers on eBay are more influential than moderate ratings. In contrast, \citet{Mudambi.2010} showed that for electronic devices, extreme reviews are less helpful than moderate reviews. 
In this paper, we hypothesize that a potential reason for these contradictory findings is that they study two-sidedness solely in the context of review ratings, whereas a review's actual textual content is ignored.
Based on this notion, we derive our research hypotheses, all of which aim at studying two-sidedness in reviews from a language-based perspective. 


The line of arguments and their reasoning plays a key role regarding the interpretation of information. For example, \citet{Tversky.1974} find that the increased availability of justifications for a decision increases the confidence of the decision-maker. Similarly, \citet{Schwenk.1986} shows that the arguments of managers are more persuasive when they provide more information in support of their position. This preference for diagnostic information can be based on multiple factors. For instance, a person may not yet have made the cognitive effort to identify the reasons for a decision. Similarly, the person might not be motivated to weigh the pros and cons regarding various alternatives. Hence, in the context of reviews, one could expect that an in-depth review from someone who has already expended the effort to assess a product helps other customers make the purchase decision. Reviews provide detailed information when presenting a balanced, two-sided view of both the pros and cons. We thus expect reviews with a higher degree of two-sidedness to exhibit a greater perceived helpfulness as compared to one-sided appeals with a clear-cut positive or negative opinion. Therefore, our first research hypothesis states:

\vspace{.3em}
\noindent
\emph{\textbf{Hypothesis~1.}} \textit{A higher degree of two-sidedness increases the helpfulness of a review.}
\vspace{.3em}


Existing literature has produced mixed results regarding the question of whether positive or negative reviews are more helpful to customers. A possible reason for the inconsistent findings in previous works is that they ignore the initial beliefs of customers before assessing a product review \citep{Yin.2016}. 
Consumers evaluate reviews from other customers in order to help them fulfill their consumption goals \citep{Zhang.2010}. Positive reviews provide information about satisfactory experiences with the product, and thus represent opportunities to attain positive outcomes. Since positive reviews are more congruent with consumers' goals, they are likely to be more persuasive than negative ones (positivity bias). Therefore, we may expect that a positive review refuting negative arguments may remove the lingering doubts of a customer and provide particularly convincing information in his or her decision process \citep{Pan.2011}. We thus expect the role of two-sidedness to be stronger for positive reviews as compared to negative reviews. Therefore, H2 states:


\vspace{.3em}
\noindent
\emph{\textbf{Hypothesis~2.}} \textit{The effect of two-sidedness on review helpfulness is stronger for positive reviews.}
\vspace{.3em}


Another important question is how the dispersion of review ratings influences the effect of two-sided argumentation on helpfulness. A high dispersion of ratings indicates low agreement among reviewers, who exhibit a range of diverging opinions about a product. In addition, a higher rating dispersion indicates a higher relevancy of diversity in customers' tastes or product details \citep{Clemons.2006}. We thus expect two-sidedness to be particularly informative when the dispersion of ratings is high. Thus, H3 states:

\vspace{.3em}
\noindent
\textbf{\emph{\textbf{Hypothesis~3.}}} \textit{A higher rating dispersion increases the effect of two-sidedness on review helpfulness.}
\vspace{.3em}


Besides a cognitive thinking dimension, human perception is also influenced by an affective feeling dimension \citep{Sweeny.2010}. Similar to other textual information sources, product reviews can be highly emotionally charged. Such personal content cannot assumed to be uniformly helpful to the purchase decision. In contrast, customers are more likely to seek objective, factual information that contains information about how the product is used and how it compares to alternatives \citep{Ghose.2007}. Emotionally charged messages, however, typically strengthen opinions in a one-sided direction and have the potential to distract from relevant factual information \citep{Prollochs.2016,Sweeny.2010}. Therefore, we expect a higher degree of emotionality to decrease the effect of two-sidedness on review helpfulness. H4 states:

\vspace{.3em}
\noindent
\emph{\textbf{Hypothesis~4.}} \textit{A higher degree of emotionality decreases the effect of two-sidedness on review helpfulness.}

\section{Methodology}
\label{sec:methodology}

This section introduces our methodology by which to infer the degree of two-sidedness of argumentation in customer reviews. For this purpose, we employ a two-staged approach: First, the review texts are mapped to a vector-based representation using sentence embeddings. We then combine the vector representations with the review ratings to infer the polarity of individual sentences using multi-instance learning.

\subsection{Distributed Sentence Representations} 


The accuracy of textual analysis depends heavily on the representation of the textual data and the selection of features \citep{Le.2014,Prollochs.2018}. To overcome the drawbacks of the frequently employed bag-of-words approach, such as missing context and information loss, we take advantage of recent advances in learning distributed representations for text. 
For this purpose, we employ the \textit{doc2vec} library developed by Google \citep{Le.2014}. This library is based on a deep learning model that creates numerical representations of texts, regardless of their length. Specifically, the underlying model allows one to create distributed representations of sentences by mapping the textual data onto a vector space. The resulting sentence vectors have several useful properties. First, more similar sentences are mapped to more similar vectors. 
Second, the feature vectors also fulfill simple algebraic properties such as, for example, \emph{king} - \emph{man} + \emph{woman} = \emph{queen}. 
The feature representations created by the \textit{doc2vec} library have been shown to significantly increase the accuracy of text classification \citep{Le.2014}. 


For the training of our \emph{doc2vec} model, we initialize the word vectors with the vectors from the pre-trained Google News dataset\footnote{Available from the Google code archive at \url{https://code.google.com/archive/p/word2vec/}.}, which is the predominant choice in the previous literature. 
Here we use the hyperparameter settings developed by \citet{Lau.2016} during an extensive analysis. 
Subsequently, we split each review into sentences and generate vector representations for all sentences.
These are used in the next section as input data to infer polarity labels for individual sentences using multi-instance learning. 

\subsection{Inferring Two-Sidedness of Argumentation Using Multi-Instance Learning}


We are facing a problem in which the observations (reviews) contain groups of instances (sentences) instead of a single feature vector, whereby each review is associated with a rating. Formally, let $X = \{\boldsymbol{x}_i\}, i=1\dots N$ denote the set of all sentences in all reviews, $N$ the number of sentences, $D$ the set of reviews and $K$ the number of reviews. Each review $D_k=(\mathcal{G}_k, l_k)$ consists of a multiset of sentences $\mathcal{G}_k \subseteq X$ and is assigned a label $l_k$ ($0$ for negative and $1$ for positive).  The learning task is to train a classifier $y$ with parameters $\boldsymbol{\theta}$ to infer sentence labels $y_{\boldsymbol{\theta}}(\boldsymbol{x}_i)$ given only the review labels.

The above problem is a multi-instance learning problem~\citep{Dietterich.1997}, which can be solved by constructing a loss function consisting of two components: (a)~a term that punishes different labels for similar sentences; (b)~a term that punishes misclassifications at the review level. 
The loss function $L(\boldsymbol{\theta})$ is then minimized as a function of the classifier parameters $\boldsymbol{\theta}$, 

\vspace{-0.6cm}
\small
\begin{align}
L(\boldsymbol{\theta}) &= \frac{1}{N^2} \sum\limits_{i=1}^N \sum\limits_{j=1}^N \mathcal{S}(\boldsymbol{x}_i,\boldsymbol{x}_j) (y_i - y_j)^2 + \frac{\lambda}{K} \sum\limits_{k=1}^K (A(D_k,\boldsymbol{\theta}) - l_k)^2,  \label{eq:costgeneral}
\end{align}
\vspace{-0.4cm}
\normalsize

where $\lambda$ is a free parameter that denotes the contribution of the review level error to the loss function. In this function, $\mathcal{S}(\boldsymbol{x}_i,\boldsymbol{x}_j)$ measures the similarity between two sentences $\boldsymbol{x}_i$ and $\boldsymbol{x}_j$, and $(y_i - y_j)^2$ denotes the square loss on the predictions for sentences $i$ and $j$. In addition, $A(D_k,\boldsymbol{\theta})$ denotes the predicted label for the review $D_k$. Hence, the loss function punishes different labels for similar sentences while still accounting for a correct classification of the review label.
In order to adapt the loss function to our problem, \ie classifying sentences in reviews into positive and negative categories, we specify concrete functions for the placeholders in Equation \ref{eq:costgeneral} as follows. First, we use cosine similarity to calculate a similarity measure between two sentence representations, \ie  $\mathcal{S}(\boldsymbol{x}_i,\boldsymbol{x}_j) = \frac{\boldsymbol{x}_i \cdot \boldsymbol{x}_j}{||\boldsymbol{x}_i|| \cdot ||\boldsymbol{x}_j||}$. Second, we need to specify a classifier to predict $y_i$. Here, we choose a logistic regression model due to its simplicity and reliability. 
Third, 
we define $A(D_k,\boldsymbol{\theta})$ as the most frequent label of the sentences $\mathcal{G}_k$.
Altogether, this results in a specific loss function which is to be minimized by the parameter of the logistic regression $\boldsymbol{\theta}$ using stochastic gradient descent.

Ultimately, we use the above model to infer labels of individual sentences as follows. First, a sentence is transformed into its vector representation $\boldsymbol{x}_i$. Second, we calculate $y_{\boldsymbol{\theta}}(\boldsymbol{x}_i)$ via the logistic regression model. If the result of $y_{\boldsymbol{\theta}}(\boldsymbol{x}_i)$ is greater than or equal to \SI{0.5}, the model predicts positive (and negative otherwise). It is worth noting that this approach yields \SI{90.30}{\percent} accuracy on a manually-labeled, out-of-sample dataset of \num{1000} positive and negative sentences from Amazon reviews, which can be seen as reasonably accurate for our analysis. 
In contrast to alternative approaches, such as dictionary-based methods or supervised learning models, the method yields superior performance and does not require any kind of manual labeling. 
Based on the sentence polarity labels, we then determine the degree of two-sidedness $RTS$ in a review $D_k$ via 

\vspace{-0.6cm}
\small
\begin{align}
RTS = 1- \left|\frac{1}{|\mathcal{G}_k|}  \sum_{x_i \in \mathcal{G}_k}y_{\boldsymbol{\theta}}(\boldsymbol{x}_i) -0.5 \right| \cdot 2.
\end{align}
\vspace{-0.4cm}
\normalsize

Hence, we map reviews with an equal number of positive and negative sentences to the value $1$ and reviews with either only positive or only negative sentences to the value $0$.\footnote{{For reasons of simplicity and reproducibility, we follow previous literature by classifying sentences into positive and negative categories. As a robustness check, we also tested an alternative variant with an additional neutral category. This approach yields a similar distribution for $RTS$ and qualitatively identical results in our later analysis.}}

\section{Empirical Analysis}
\label{sec:results}


\subsection{Dataset and Empirical Model}


For our analysis, we use a frequently-employed corpus of retailer-hosted consumer reviews in the category of cell phones and accessories from Amazon \citep{He.2016}. This dataset exhibits several advantages as compared to alternative sources: first, all reviews are verified by the retailer, \ie the author of a review must have actually purchased the product. Second, the Amazon platform features a particularly active user base, \ie a high number of reviews per product \citep{Gu.2012}.
The complete sample consists of \num{192189} consumer reviews containing the following information: (1) the numerical rating assigned to the product (\ie the star rating), (2) the review text, (3) the number of helpful votes for the review, (4) the date on which the review was posted. Moreover, we collected the following product-specific information: (i) the price of the product, and (ii) the average star rating. 
In addition, we determine the two-sidedness for all reviews in our dataset using the methodology described in the previous section. This measure ranges from 0 (only one-sided arguments) to 1 (an equal number of positive and negative sentences). Out of all reviews, \SI{66.56}{\percent} contain both positive and negative sentences. A share of \SI{22.19}{\percent} of all documents contain only positive sentences, while \SI{11.25}{\percent} consist solely of negative sentences. The mean two-sidedness in our dataset is \num{0.54}. The average number of sentences per review is \num{6.15}. 
The average star rating of a product is \num{4.13}. Reviews have received helpful votes in a range between \num{0} and \num{158}. The mean number of helpful votes is \num{1.04}. The mean length of a review is \num{94.85} words. 

We use a quasi-poisson model to analyze the effect of two-sided argumentation on the perceived helpfulness of customer reviews. This type of model is not only a common choice for the analysis of word-of-mouth variables, but also has the advantage of being able to handle the many count variables in our dataset. The dependent variable in the model is the helpfulness of a review, given by the number of helpful votes from other users $RHVotes$. 
The key explanatory variable is $RTS$, which measures the degree of two-sidedness in a review. 
Consistent with the related literature \cite[\eg][]{Yin.2016}, we additionally use a fixed set of control variables for each product, namely, a product's average rating ($PAvg$), the dispersion of ratings ($PDisp$) and the price ($PPrice$). 
In addition, we incorporate the following control variables at the review level \citep{Lutz.2017, Mudambi.2010, Yin.2016}. We include the review age in years ($RAge$), the review length in increments of 100 words ($RLength$) and the difference between the review rating and the product's average rating ($RDiff$). In addition, we add a variable $REmo$ that allows us to measure the degree of emotionality of a review. The emotionality measure is calculated based on the fraction of emotional words in a review using the frequently-employed NRC dictionary \citep{Mohammad.2010}. Following prior research, we also control for the fraction of cognitive words in a review ($RCog$) using the LIWC text analysis software \citep{Yin.2016}.
Altogether, the resulting model with intercept $\beta_0$ and error term $\epsilon$ is

\vspace{-.6cm}
\begin{flalign}
	Ln(RHVotes) &= \beta_0 
	+ \beta_1 \,\mathit{PAvg}
	+ \beta_2 \,\mathit{PDisp}
	+ \beta_3 \,\mathit{PPrice}
	+ \beta_4 \,\mathit{RAge}
	+ \beta_5 \,\mathit{RCog}
	+ \beta_6 \,\mathit{REmo}
	+ \beta_7 \,\mathit{RDiff} \\
	&+ \beta_8 \,\mathit{RLength}
	+ \beta_9 \,\mathit{RTS}
	+ \beta_{10} \,\mathit{RTS} \times \mathit{RDiff}
	+ \beta_{11} \,\mathit{RTS} \times \mathit{PDisp} \nonumber
	+\beta_{12} \, \mathit{RTS} \times \mathit{REmo}  \nonumber
	+ \epsilon. &&
	\label{eqn:regression_h3}
\end{flalign}
\vspace{-.6cm}

\subsection{Hypotheses Tests}

We now use the above model to test our hypotheses. All regression results are provided in \Cref{tbl:results}. We start our analysis with a baseline model in which we only include the independent variables from previous works. The results are shown in column (a) of \Cref{tbl:results}. The analysis of the model indicates a good fit, with a relatively high McFadden's pseudo $R^2$ value of \num{0.2086}. As expected, the length and age of a review have a positive impact on review helpfulness. In contrast, a higher fraction of cognitive words, a higher degree of emotional language and a higher difference between rating and average rating have a negative impact. We also see that more expensive products tend to have more helpful reviews.

We now test our first hypothesis. For this purpose, we add the variable $RTS$ to our model. The results are shown in column (b) of \Cref{tbl:results}. The coefficient of $RTS$ is significant and positive ($\beta=0.864, p < 0.001$). Hence, more two-sided reviews containing positive as well as negative arguments exhibit a greater helpfulness for other customers. We also note an increase in terms of $R^2$ from \num{0.2086} for the baseline model to \num{0.2238}. All other coefficients remain stable. Thus, H1 is supported.

To test our second hypothesis, we extend our previous model by additionally adding the interaction term $RTS \times RDiff$. The results for this model are shown in column (c). The coefficient of this interaction is positive and significant ($\beta = 0.185, p < 0.001$). Hence, we find support for H2 stating that the effect of two-sided argumentation is higher for positive reviews \ie for reviews with a rating above a product's average rating. Next, we test whether a higher degree of rating dispersion increases the effect of two-sided argumentation on helpfulness. For this purpose, we modify the model in column (c) to additionally include the interaction term $RTS \times PDisp$. The results of this model are shown in column (d). The additional term is positive but not statistically significant at any common significance level. Thus, we do not find support for H3. Finally, we test our fourth hypothesis by adding another interaction term $RTS \times REmo$. The coefficient of this interaction term is negative and significant ($\beta = -1.324, p < 0.001$). Thus, we find support for H4 stating that a higher degree of emotionality decreases the effect of two-sided argumentation.


Ultimately, we perform several robustness checks to prove the validity of our analysis. First, we check our models for possible multicollinearity. For this purpose, we calculate the variance inflation factors (VIF) for all variables in our models. The VIF of all regressors (except the interaction terms) are below the critical threshold of 4. This finding is also supported by the fact that our independent variables show relatively high significance values with comparatively low standard errors. 
Second, we also validate our model by adding quadratic terms of $RTS$ to the individual models. According to our results, the additional terms are not statistically significant and all models continue to support our hypotheses. Third, we tested the extend to which the emotionality measure based on the NRC emotions dictionary also reflects the subjectivity of a review. For this purpose, we tested an alternative model in which we replaced $REmo$ with a corresponding subjectivity measure based on the MPQA subjectivity lexicon \citep{Wilson.2005}. The models yield similar results and a significantly negative effect for subjectivity on review helpfulness. Hence, to a certain extent, the emotionality measure also reflects the subjectivity of a review.

\newcommand{\numObs}{\multicolumn{1}{c|}{$192,189$}}

\begin{table}
\renewcommand{\arraystretch}{1}
\begin{center}
\small
\begin{tabular}{|l| D{.}{.}{4.7}| D{.}{.}{4.7}| D{.}{.}{4.7}| D{.}{.}{4.7}| D{.}{.}{4.7}|}
\hline
\multicolumn{6}{|c|}{\rule[-.4cm]{0pt}{1cm}\textbf{\normalsize\Cref{tbl:results}. Regression Linking Two-Sidednessness and Review Helpfulness. }}   \\ \hline
 & \multicolumn{1}{c|}{\textbf{(a)}} & \multicolumn{1}{c|}{\textbf{(b)}} & \multicolumn{1}{c|}{\textbf{(c)}} & \multicolumn{1}{c|}{\textbf{(d)}}  & \multicolumn{1}{c|}{\textbf{(e)}}\\
\hline
 $Intercept$ & -2.030^{***} & -2.551^{***} & -2.550^{***} & -2.545^{***} & -2.630^{***} \\ 
  & (0.184) & (0.181) & (0.181) & (0.191) & (0.192) \\ 
  \hline
  $PAvg$& 0.042 & 0.073^{*} & 0.075^{*} & 0.075^{*} & 0.076^{*} \\ 
  & (0.032) & (0.031) & (0.031) & (0.031) & (0.031) \\ 
  \hline
  $PDisp$ & 0.00002 & 0.070 & 0.058 & 0.054 & 0.053 \\ 
  & (0.045) & (0.044) & (0.044) & (0.069) & (0.069) \\ 
  \hline
 $PPrice$ & 0.380^{***} & 0.353^{***} & 0.355^{***} & 0.355^{***} & 0.353^{***} \\ 
  & (0.012) & (0.012) & (0.012) & (0.012) & (0.012) \\ 
  \hline
 $RAge$& 0.389^{***} & 0.373^{***} & 0.374^{***} & 0.373^{***} & 0.373^{***} \\ 
  & (0.005) & (0.005) & (0.005) & (0.005) & (0.005) \\ 
	  \hline
 $RCog$ & -1.224^{***} & -1.724^{***} & -1.844^{***} & -1.843^{***} & -1.839^{***} \\ 
  & (0.220) & (0.224) & (0.224) & (0.224) & (0.224) \\ 
  \hline
 $REmo$ & -2.500^{***} & -2.449^{***} & -2.464^{***} & -2.464^{***} & -1.784^{***} \\ 
  & (0.135) & (0.136) & (0.136) & (0.136) & (0.220) \\ 
  \hline
$RLength$  & 0.122^{***} & 0.114^{***} & 0.114^{***} & 0.114^{***} & 0.113^{***} \\ 
  & (0.001) & (0.001) & (0.001) & (0.001) & (0.001) \\ 
  \hline
 $RDiff$ & -0.052^{***} & -0.053^{***} & -0.159^{***} & -0.159^{***} & -0.159^{***} \\ 
  & (0.009) & (0.009) & (0.016) & (0.016) & (0.016) \\ 
  \hline
  $RTS$ &  & 0.864^{***} & 0.879^{***} & 0.871^{***} & 1.031^{***} \\ 
  &  & (0.030) & (0.030) & (0.103) & (0.111) \\ 
  \hline
  $RTS \times RDiff$ &  &  & 0.185^{***} & 0.185^{***} & 0.187^{***} \\ 
  &  &  & (0.024) & (0.024) & (0.024) \\ 
  \hline
  $RTS \times PDisp$ &  &  &  & 0.007 & 0.009 \\ 
  &  &  &  & (0.089) & (0.088) \\ 
  \hline
  $RTS \times REmo$ &  &  &  &  & -1.324^{***} \\ 
  &  &  &  &  & (0.345) \\ 
  \hline 
Observations & \numObs & \numObs & \numObs & \numObs & \numObs \\
McFadden's $R^2$ & 0.2086 & 0.2238 & 0.2248 & 0.2248 & 0.2251 \\ \hline
\multicolumn{3}{|l}{Stated: coefficient and std.~dev. in parentheses}  & \multicolumn{3}{r|}{Signif.:~$^{*}$p$<$0.05; $^{**}$p$<$0.01; $^{***}$p$<$0.001} \\ 
\hline
\end{tabular}
\end{center}
\captionsetup{labelformat=empty}
 \caption{} 
 \label{tbl:results}
\end{table}

\section{Discussion}
\label{sec:discussion}

Our study allows for a deeper understanding of the assessment of consumer reviews on online retailer platforms. In contrast to previous works that study helpfulness on the basis of structured data (such as star ratings or review length), our analysis additionally incorporates the textual dimension of customer reviews. As our main finding, we provide strong evidence that the line of arguments and their reasoning plays a key role in the interpretation of reviews. Specifically, we find that a higher degree of two-sided argumentation increases the helpfulness of a review for other users as compared to one-sided appeals with a clear-cut positive or negative opinion. This is also concordant with marketing research suggesting that two-sided messages generate a higher level of attention \citep{Crowley.1994}, as well as the experimental results from \citet{Jensen.2013}, which suggest that highlighting positive and negative aspects of a product increases the credibility of a reviewer. However, our study not only extends these works from a field study perspective, but also sheds additional light on the unresolved question of whether positive or negative reviews are more helpful to customers. In this domain, our analysis reveals an important role of two-sidedness that is stronger for positive reviews than for negative reviews. Ultimately, our analysis also indicates that the effect of two-sidedness depends on the emotional orientation of documents. 
In this respect, we see that customers prefer diagnostic, factual information about the pros and cons of a product when assessing customer reviews.

This study has implications for practitioners in the fields of marketing and public relations. Since the helpfulness of reviews is directly related to the two-sidedness of argumentation, our findings can help companies to enhance their communication strategies with regard to product descriptions, social media content, and advertisement. In this context, it should not be assumed that positive or negative reviews are generally perceived as more helpful. Instead, the role of review ratings in relation to perceived helpfulness rather depends on the line of arguments presented in the review. In a next step, our findings can also help retailer platforms to better inform customers who are considering purchasing a product. For instance, retailer platforms might utilize our findings to develop writing guidelines to encourage more useful seller reviews. It is worth noting that a better understanding of why customers perceive a particular review as helpful or unhelpful can also aid in the detection of fake reviews \citep{Zhang.2016}.

\section{Conclusion and Further Research}
\label{sec:conclusion}

A growing body of literature is attempting to clarify the influence of word-of-mouth on customer purchase decisions. 
In this paper, we examine the effect of two-sided argumentation on the perceived helpfulness of Amazon customer reviews. In contrast to previous works, our analysis thereby sheds light on the reception of reviews from a language-based perspective. According to our results, two-sided argumentation in reviews significantly increases their helpfulness. We find this effect to be stronger for positive reviews than for negative reviews, whereas a higher degree of emotional language weakens this effect. In a practical sense, our results allow practitioners in the fields of marketing and public relations to enhance their communication strategies. 
Moreover, we contribute to IS research by addressing the question of how textual information affects customers' individual behavior and decision-making.

On the road to completing this research in progress, we will expand the study in four directions. First, our dataset is limited to reviews about cell phones and accessories. To analyze the generalizability of our results, we will examine the differential impact of two-sidedness of argumentation in the context low-involvement and high-involvement products. 
Second, it might be interesting to analyze the effects of two-sidedness on other recommendation platforms, such as restaurant reviews or social media. Third, it is an intriguing notion to study how two-sidedness and its relevancy depends on the coverage of different aspects and topics in reviews. Fourth, further research is necessary to study the differences in information reception among different target groups. For instance, 
the subjective interpretation of the same information might vary across different audiences and cultures.

\bibliographystyle{misqNic}
\bibliography{bib/literature}
\end{document}